\documentclass[runningheads,envcountsame,oribibl]{llncs}
\usepackage[colorinlistoftodos,prependcaption,textsize=tiny]{todonotes}

\usepackage{amssymb}
\usepackage{amsfonts}
\usepackage{amsmath}
\usepackage{dsfont} 
\usepackage{graphicx}
\usepackage{lineno}
\usepackage{rotating}
\makeatletter
\newcommand{\figcaption}{\def\@captype{figure}\caption}
\newcommand{\tabcaption}{\def\@captype{table}\caption}
\makeatother

\usepackage[caption=false]{subfig}

\usepackage[misc]{ifsym}  
\usepackage{booktabs}
\usepackage{ctable}
\newcolumntype{+}{>{\global\let\currentrowstyle\relax}}
\newcolumntype{^}{>{\currentrowstyle}}

\newcommand{\PreserveBackslash}[1]{\let\temp=\\#1\let\\=\temp}
\newcolumntype{C}[1]{>{\PreserveBackslash\centering}p{#1}}
\newcolumntype{R}[1]{>{\PreserveBackslash\raggedleft}p{#1}}
\newcolumntype{L}[1]{>{\PreserveBackslash\raggedright}p{#1}}

\usepackage{makecell}
\usepackage{multirow}

\usepackage{microtype}
\usepackage[colorlinks=true,citecolor=blue,urlcolor=black]{hyperref}
\usepackage[capitalize]{cleveref}

\newcommand{\R}[0]{\mathds{R}} 
\newcommand{\C}[0]{\mathds{C}} 

\providecommand{\mb}[1]{\mathbf{#1}}

\usepackage{soul}

\usepackage{mathtools}
\newtagform{myform}{\normalsize(}{)}
\usetagform{myform}

\usepackage{microtype}
\usepackage[colorlinks=true,citecolor=blue,urlcolor=black]{hyperref}

\usepackage{adjustbox}

\newcommand\blfootnote[1]{%
  \begingroup
  \renewcommand\thefootnote{}\footnote{#1}%
  \addtocounter{footnote}{-1}%
  \endgroup
}

\begin{document}

\title{dAUTOMAP: decomposing AUTOMAP to achieve scalability and enhance performance }

\titlerunning{ISMRM Abstract \#658: dAUTOMAP}  

\newcommand{\corrauth}{\textsuperscript{(\Letter)}}
\author{Jo Schlemper\inst{1}\corrauth \and Ilkay Oksuz\inst{2} \and James Clough\inst{2} \and Jinming Duan\inst{1} \and Andrew P. King\inst{2} \and Julia A. Schnabel \and \inst{2} Jo V. Hajnal\inst{3} \and Daniel Rueckert\inst{1}}
\authorrunning{J. Schlemper et al.}
\institute{Biomedical Image Analysis Group, Imperial College London, UK \email{\{jo.schlemper11,d.rueckert\}@imperial.ac.uk}
\and School of Biomedical Engineering and Imaging Sciences, King's College London, UK\\
\email{\{o.oksuz,j.clough,andrew.king,julia.schnabel\}@kcl.ac.uk}
\and Imaging and Biomedical Engineering Clinical Academic Group, King's College London, UK\\ \email{\{jo.hajnal\}@kcl.ac.uk}
}

\mainmatter
\maketitle              

\begin{abstract}
 AUTOMAP \cite{zhu2018image} is a promising generalized reconstruction approach, however, it is not scalable and hence the practicality is limited. We present dAUTOMAP, a novel way for decomposing the domain transformation of AUTOMAP, making the model scale linearly. We show dAUTOMAP outperforms AUTOMAP with significantly fewer parameters.
\end{abstract}

\section{Introduction}

Recently, automated transform by manifold approximation (AUTOMAP) \cite{zhu2018image} has been proposed as an innovative approach to directly learn the transformation from source signal domain to target image domain. While the applicability of AUTOMAP to a range of tasks has been demonstrated, its practicality remains limited as the required number of parameters scales quadratically with the input size. We present a novel way for decomposing the domain transformation, which makes the model scale linearly with the input size. We term the resulting network \emph{dAUTOMAP} (decomposed - AUTOMAP). We show that, remarkably, the proposed approach outperforms AUTOMAP for the provided dataset with significantly fewer parameters.\blfootnote{Presented at ISMRM 27th Annual Meeting \& Exhibition (Abstract \#658)}

\section{Methods}
Let $\mb{x} \in \C^{N\!\times\!M}$ be a complex-valued image. The two-dimensional Discrete Fourier Transform (DFT) is given by:
\begin{align}
\mb{y}[k,l] = \sum_{n=0}^{N} \sum_{m=0}^{M} \mb{x}[n, m] e^{-j2\pi (\frac{nk}{N} + \frac{ml}{M})}
\end{align}
This is commonly written as a matrix product: $\text{vec}(\mb{y}) = \mb{E}\text{vec}(\mb{x})$, where we take row-major vectorization and $\mb{E} = e^{-j2\pi (\frac{nk}{N} + \frac{ml}{M})}$, with $p = kM+l$, $q=nM+m$. As the matrix $\mb{E}$ is the Kronecker product of two one-dimensional DFT's, we have:
\begin{align}
    \mb{E} \text{vec}(\mb{x}) =
    \left( \mb{F}_N \otimes \mb{F}_M \right) \text{vec}(\mb{x}) = 
    \text{vec} \left(\mb{F}_N\mb{x}\mb{F}_M^H \right) = 
    \text{vec} \left( (\mb{F}_M(\mb{F}_N\mb{x})^H)^H \right),
\end{align}
where $(\mb{F}_N)_{kn} =  e^{-j2\pi \frac{nk}{N}}$,  $(\mb{F}_M)_{lm} =  e^{-j2\pi \frac{ml}{M}}$. Observe that $\mb{F}_N \mb{x}$ can be computed using a convolution layer with $N$ kernels of size $(N,1)$ with no padding where the output tensor has size $(N_\text{batch}, N, 1, M)$. Motivated by this, we propose a decomposed transform layer (DT layer): a convolution layer with the above kernel size, which is learnable. In the simplest case, the layer can be reduced to the (inverse) Fourier transform or identity. A 2D DFT can be performed by applying the DT layer twice, where the intermediate tensor is first reshaped into $(N_\text{batch}, 1, N, M)$ and then conjugate-transposed. Note that the complex nature of the operation is preserved by $\R^2$, which doubles the number of output channels (i.e. $2N$). Therefore, the convolution kernel of the DT layer has the shape: $(N_{c_\text{out}}, N_{c_\text{in}}, \text{kernel}_x, \text{kernel}_y) = (2N, 2, N, 1)$. For the second DT layer, $N$ and $M$ are swapped.

\begin{figure}[t]
    \centering
    \includegraphics[width=\textwidth]{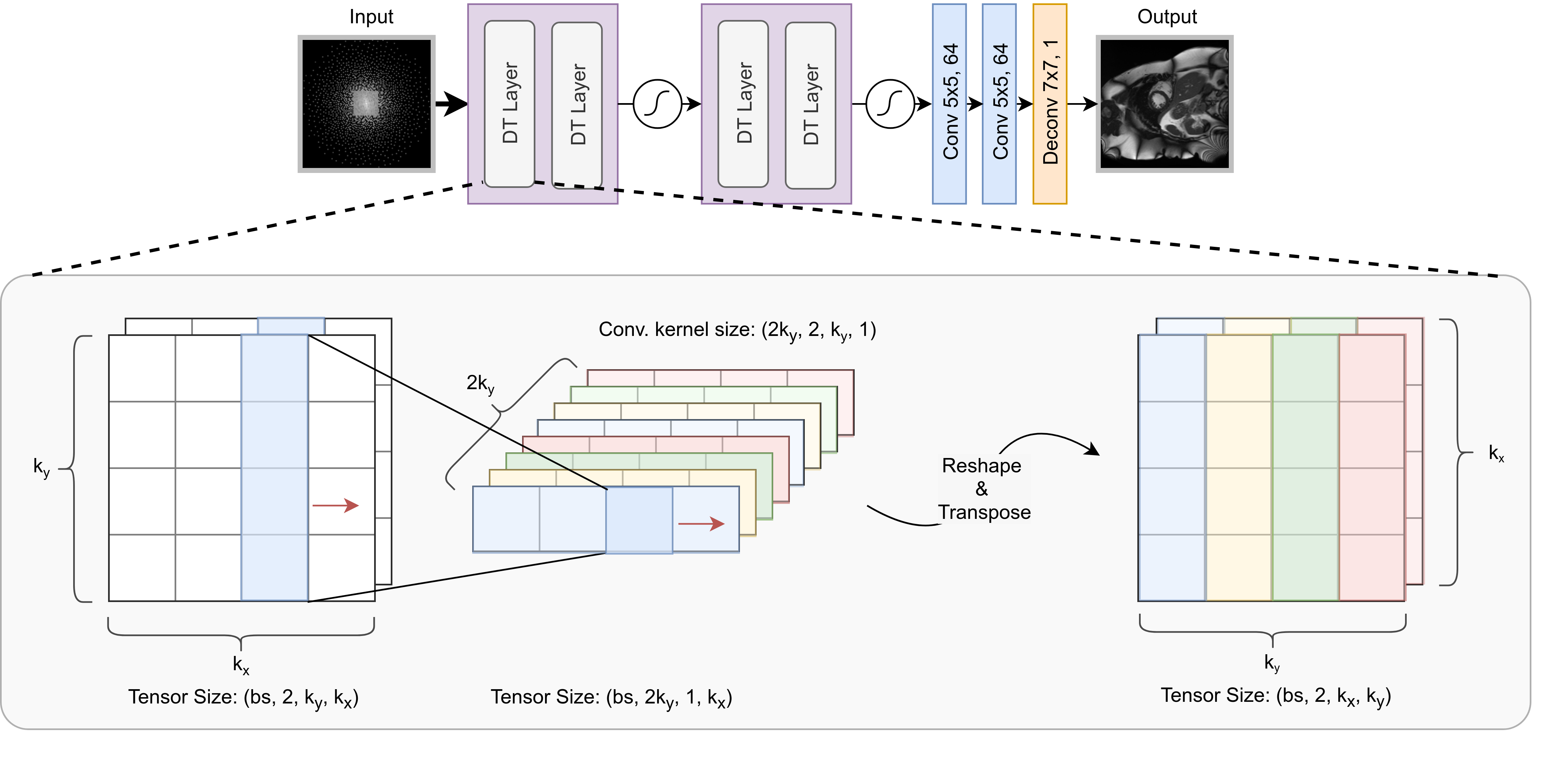}
    \caption{The proposed network architecture of dAUTOMAP. The network takes $k$-space data on a Cartesian grid and directly recovers grid and directly recovers the underlying image. The fully-connected layers in AUTOMAP are replaced by decomposed transform (DT) layers, which are fully-connected along one axis. For two-dimensional input, we apply the DT layer twice, once for each axis. Each DT block is activated by ReLU nonlinearity. DT blocks are followed by a sparse convolutional autoencoder, as proposed in \cite{zhu2018image}.}
    \label{fig:dautomap_architecture}
\end{figure}

The proposed dAUTOMAP, shown in \cref{fig:dautomap_architecture}, replaces the fully-connected layers in AUTOMAP by DT layers. We used ReLU as the choice of non-linearity.

\section{Evaluation} 

We evaluated the proposed method on a simulation-based study using short-axis (SA) cardiac cine magnitude images from the UK Biobank Study\cite{petersen2015uk} ($\!>\!$1M SA slices). To compare with AUTOMAP, the data were subsampled to central $128 \!\!\times\!\! 128$ k-space grid points. Both methods were evaluated on the reconstruction tasks from three undersampling patterns: (1) Cartesian with Acceleration Factor AF$=$2, (2) Poisson with AF$=$4, (3) Variable density Poisson (VDP) with AF$=$7\cite{uecker2015berkeley}. For dAUTOMAP, we also experimented with the images having $256 \!\!\times\!\! 256$ k-space grid points, with $2\times$ Cartesian undersampling.

Both networks were initialised randomly and trained for 1000 epochs. We used RMSProp with $lr\!\!=\!\!2 \!\!\times\!\! 10^{-5}$ and Adam optimiser $lr\!\!=\!\!10^{-3}$ for AUTOMAP and dAUTOMAP respectively. The reconstructions were evaluated by mean squared error (MSE), peak signal-to-noise ratio (PSNR), structural similarity (SSIM) and High Frequency Error Norm (HFEN). We also compared the reconstruction speed and the required parameters. 

\section{Results}


\begin{table}[!t]
\centering
\caption{Quantitative comparison between AUTOMAP and dAUTOMAP. In general, dAUTOMAP outperformed AUTOMAP for Mean Squared Error (MSE), peak signal-to-noise ratio (PSNR), structural similarity (SSIM) and high frequency error norm (HFEN).} 
\label{table:dautomap_quantitative}
\begin{adjustbox}{max width=\textwidth}
\begin{tabular}{ccccccccc}
&&&          & \multicolumn{4}{c}{}                                                       \\
$N_\text{grid}$ & AF & Undersampling & Model    & MSE ($\times 10^{-3}$)         & PSNR (dB)        & SSIM       & HFEN      \\
\toprule
\multirowcell{2}{$128$} & \multirowcell{2}{7}
& \multirowcell{2}{VDP} & AUTOMAP  & 3.27$\pm$1.17 & 22.07$\pm$1.35 & 0.76$\pm$0.03 & 0.54$\pm$0.07 \\
&&                            & dAUTOMAP & {\bf 1.76}$\pm${\bf 0.47}& {\bf 24.67}$\pm${\bf 1.14}&{\bf 0.82}$\pm${\bf 0.02}&{\bf 0.39}$\pm${\bf 0.03}\\
\cmidrule(lr){1-8}
\multirowcell{2}{$128$} & \multirowcell{2}{4}
& \multirowcell{2}{Poisson}   & AUTOMAP  & 3.63$\pm$1.28 & 21.61$\pm$1.34 & 0.74$\pm$0.03 & 0.62$\pm$0.09 \\
&&                            & dAUTOMAP & {\bf 1.54}$\pm${\bf 0.43}& {\bf 25.28}$\pm${\bf 1.20}&{\bf 0.84}$\pm${\bf 0.02}&{\bf 0.40}$\pm${\bf 0.03}\\
\cmidrule(lr){1-8}
\multirowcell{2}{$128$} & \multirowcell{2}{2}
& \multirowcell{2}{ Cartesian }     & AUTOMAP  & 2.81$\pm$1.35 & 22.84$\pm$1.64 & 0.80$\pm$0.03 & 0.42$\pm$0.07 \\
&&                            & dAUTOMAP & {\bf 1.01}$\pm${\bf 0.39}& {\bf  27.2 }$\pm${\bf 1.51}&{\bf 0.89}$\pm${\bf 0.02}&{\bf 0.27}$\pm${\bf 0.05}\\
\cmidrule(lr){1-8}
\multirowcell{2}{$256$} & \multirowcell{2}{2}
& \multirowcell{2}{Cartesian} & AUTOMAP  & n/a & n/a & n/a & n/a \\
&&                            & dAUTOMAP & {\bf 0.51}$\pm${\bf 0.25}& {\bf 30.31}$\pm${\bf 1.81}&{\bf 0.91}$\pm${\bf 0.02}&{\bf 0.29}$\pm${\bf 0.04}\\
\bottomrule
\end{tabular}
\end{adjustbox}
\end{table}

\begin{figure}[t!]
    \centering
    \includegraphics[width=\textwidth]{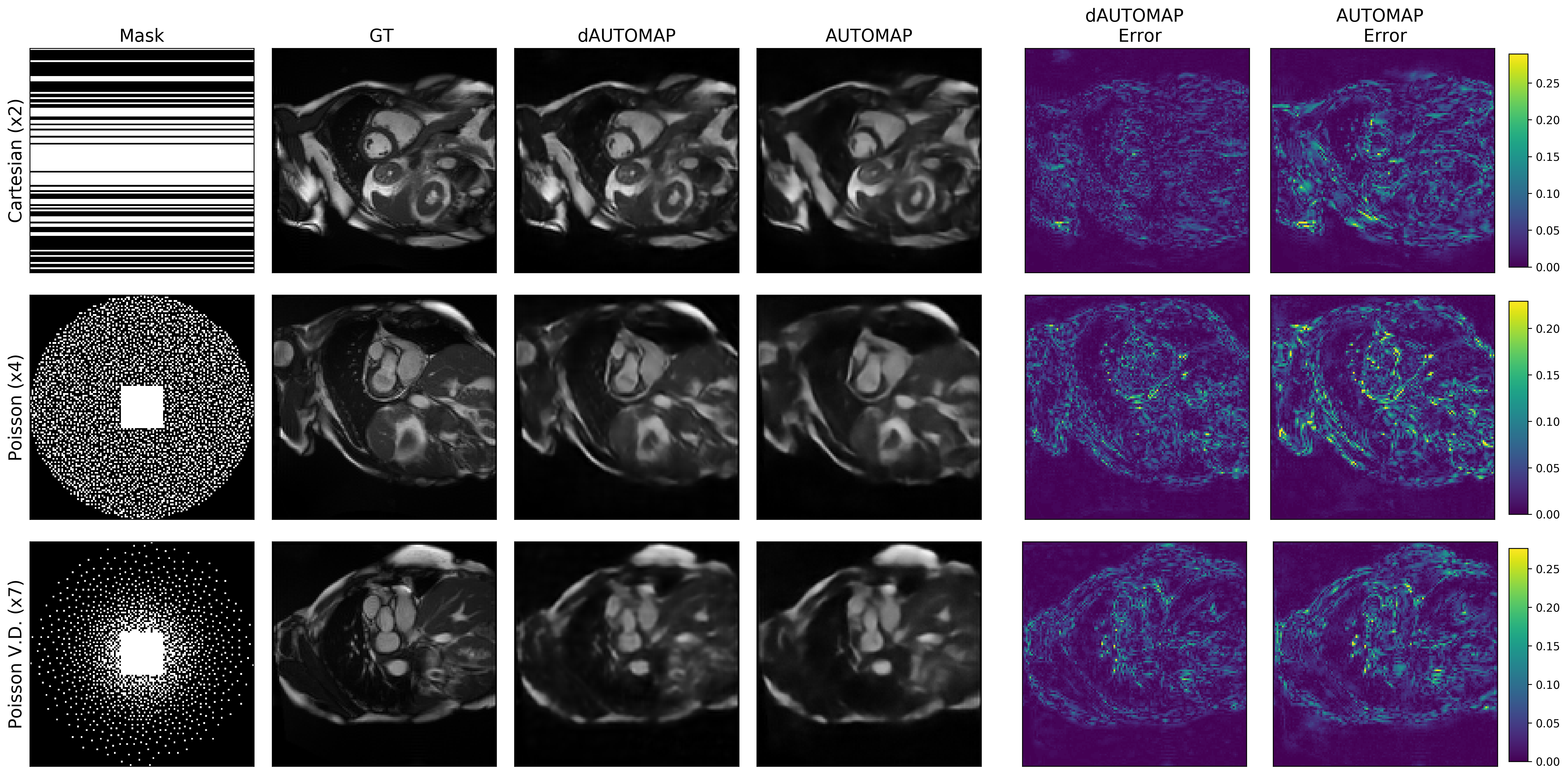}
    \caption{Reconstruction of AUTOMAP and dAUTOMAP for different undersampling patterns and the resulting error maps. One can see that AUTOMAP tends to over-smooth the edges, which were preserved by dAUTOMAP better.}
    \label{fig:qualitative_diff_sampl}
\end{figure}

\begin{figure}[t!]
    \centering
    \includegraphics[width=\textwidth]{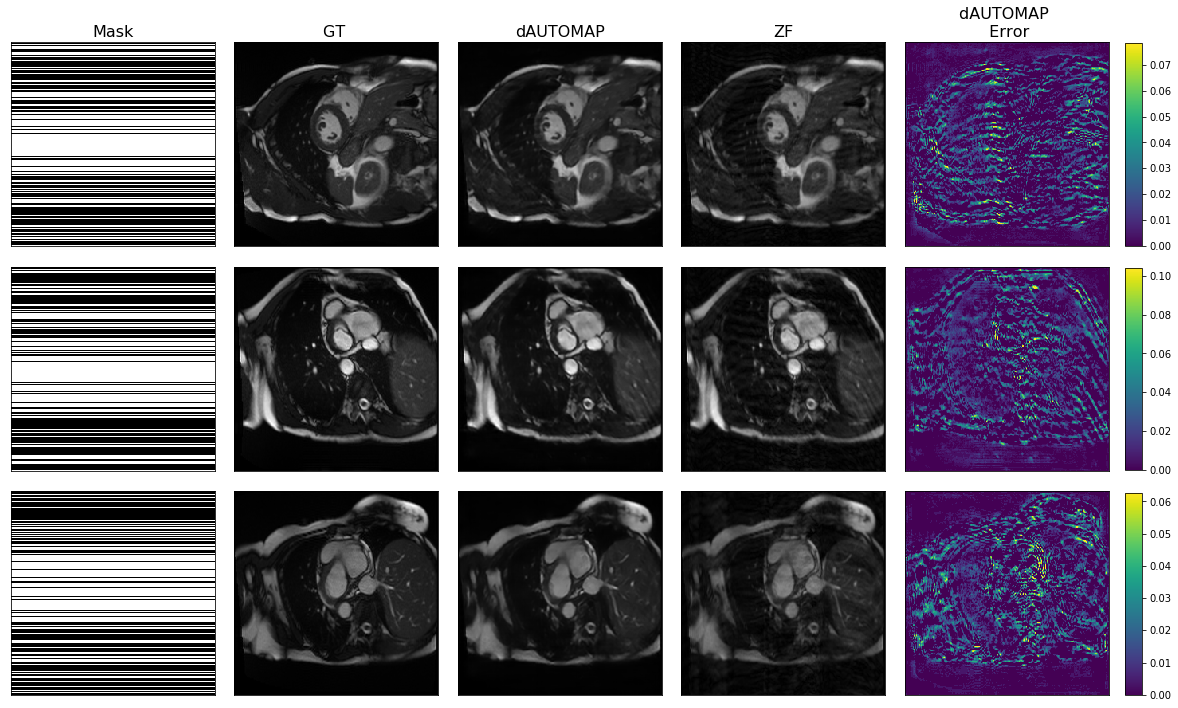}
    \caption{The sample reconstructions of dAUTOMAP from $256\!\!=\!\!256$ Cartesian $k$-space data and the corresponding error maps. The majority of the artefacts were removed, however, some high-frequency aliasing can still be observed.}
    \label{fig:qualitative_256}
\end{figure}

As shown in \cref{table:dautomap_quantitative}, the proposed approach outperformed AUTOMAP (Wilcoxon, $p \!\!\ll\!\! 0.01$). \cref{fig:qualitative_diff_sampl} shows sample reconstructions. We notice that AUTOMAP tends to over-smooth the image, whereas dAUTOMAP preserves the fine-structure better, even though the residual artefact is more prominent. The result of dAUTOMAP for $256 \!\!\times\!\! 256$ $k$-space data is shown in \cref{fig:qualitative_256}, demonstrating that the method successfully learnt a transform which simultaneously dealiases the image. The execution speeds were comparable (\cref{table:dautomap_quantitative2}. The parameters of the proposed approach required only 1.5MB of memory for $128 \!\!\times\!\! 128$ k-space data, compared to 3.1GB required for AUTOMAP (these numbers increase to 3.1MB vs 56GB for $256 \!\!\times\!\! 256$ k-space data). 

\begin{table}[t!]
\centering
\caption{Comparison of the number of parameters and execution speed.} 
\label{table:dautomap_quantitative2}
\begin{adjustbox}{max width=\textwidth}
\begin{tabular}{ccccc}
& \multicolumn{2}{c}{$N_\text{grid}\!=\!128$} & \multicolumn{2}{c}{$N_\text{grid}\!=\!256$} \\
\cmidrule(lr){2-3} \cmidrule(lr){4-5}
Model    & \#Parameters ($\!\times\!10^{6}$) & Speed (ms) & \#Parameters ($\!\times\!10^{6}$) & Speed (ms)   \\
\toprule
AUTOMAP  & 806  & 0.36$\pm$0.36 & 13000 & n/a \\ 
dAUTOMAP & 0.37 & 0.48$\pm$0.10 & 1.16 & 0.50 $\pm$ 0.12 \\
\bottomrule
\end{tabular}
\end{adjustbox}
\end{table}

\section{Discussion and Conclusion}

In this work, we propose a simple architecture which makes AUTOMAP scalable, based on the idea that the original Fourier kernels are linearly separable. We experimentally found that such an approach yields superior performance in practice, which is attributed to having significantly fewer parameters, making it easier to train and less prone to overfitting. In future, we plan to investigate the performance of non-Cartesian sampling strategies, which would require regridding, or extensions to 3D data. Finally, the code is available on \url{http://github.com/js3611/dAUTOMAP}.

\section{Acknowledgements}

Jo Schlemper is partially funded by EPSRC Grant (EP/P001009/1).

\bibliographystyle{splncs03}
\bibliography{ref}

\end{document}